\PassOptionsToPackage{numbers,sort&compress,square}{natbib}
\documentclass{article}
\usepackage[preprint]{neurips_2025}

\usepackage[utf8]{inputenc}
\usepackage[T1]{fontenc}
\usepackage[hidelinks]{hyperref}
\usepackage{url}
\usepackage{booktabs}
\usepackage{amsfonts}
\usepackage{nicefrac}
\usepackage{microtype}
\usepackage{xcolor}
\usepackage{graphicx}
\usepackage{subfig}
\usepackage{mathtools}
\usepackage{caption}
\usepackage{breakcites}

\graphicspath{{./figures/}}

\title{Synthetic Rewriting as a Quality Multiplier: Evidence from Portuguese Continued Pretraining}

\author{%
Thales Sales Almeida \\
  Institute of Computing (IC)\\
  University of Campinas (UNICAMP) \\
  Maritaca AI\\
  \And
Rodrigo Nogueira\\
  School of Electrical and Computer Engineering (FEEC)\\
University of Campinas (UNICAMP) \\
  Maritaca AI \\
  \And
Hélio Pedrini \\
  Institute of Computing (IC)\\
  University of Campinas (UNICAMP) \\
}

\begin{document}



\label{first}
\maketitle

{\fontfamily{ptm}\selectfont
\begin{abstract}
Synthetic data generation through document rewriting has emerged as a promising technique for improving language model pretraining, yet most studies focus on English and do not systematically control for the quality of the source data being rewritten. We present a controlled study of how synthetic rewriting interacts with source data quality in the context of Portuguese continued pretraining. Starting from ClassiCC-PT, a Portuguese corpus annotated with STEM and Educational quality scores, we construct two 10B-token subsets at different quality levels and rewrite each into four styles using a 7B instruction-tuned model, producing approximately 40B tokens of synthetic data per condition. We train two English-centric base models (1.1B and 7B parameters) on each condition and evaluate on PoETa V2, a comprehensive 44-task Portuguese benchmark. At the 7B scale, rewriting high-quality data yields a +3.4 NPM gain over the same data unmodified, while rewriting low-quality data provides only +0.5 NPM. At the 1.1B scale, this interaction is weaker, with unmodified low-quality data performing comparably to rewritten high-quality data. Our results demonstrate that synthetic rewriting acts primarily as a quality multiplier rather than a substitute for data curation, and that this effect is scale-dependent.
\end{abstract}
}




\section{Introduction}

Large language models (LLMs) have achieved remarkable performance gains in recent years, largely driven by increases in model scale and the availability of massive pretraining corpora. However, this progress has been uneven across languages. For many non-English languages, including Portuguese, high-quality large-scale corpora remain comparatively scarce, and training models from scratch at a competitive scale is often impractical. As a result, a common and effective strategy has been to adapt English-centric pretrained models through continued pretraining on relatively small amounts of target-language data.

Prior work has shown that continued pretraining on modest volumes of Portuguese text can substantially improve the performance of English-pretrained models across a wide range of downstream tasks, even when the additional data is several orders of magnitude smaller than the original English pretraining corpus. These findings suggest that, in data-constrained settings, careful data selection and curriculum design may be as important as raw scale. At the same time, this raises a natural question: can the gains from limited Portuguese pretraining be further amplified through data transformation, rather than data collection?

Recent studies have explored the use of synthetic data for LLM pretraining, including approaches based on paraphrasing, style transfer, and document rewriting. In particular, rewriting existing web documents into alternative styles has been shown to improve training efficiency and downstream performance, even when starting from noisy or heterogeneous sources. An important open question in this line of work concerns the role of the original data quality: to what extent can rewriting compensate for low-quality inputs, and does starting from higher-quality source material lead to systematically better models after rewriting?

In this work, we study these questions in the context of Portuguese continued pretraining under a strict data budget. We start from ClassiCC-PT~\cite{almeida2025classicc}, a large Portuguese corpus annotated with continuous STEM and Educational quality scores. Using these annotations, we construct two distinct 10B-token subsets: (i) a high-quality subset, consisting of documents with STEM or Educational scores above 2.5; and (ii) a low-quality subset, with scores between 0.5 and 2.0. This setup allows us to directly isolate the impact of original data quality under a fixed token budget.

For each subset, we apply large-scale document rewriting using a 7B-parameter instruction-tuned model, generating four rewritten versions per document following the styles proposed in prior work on web text rephrasing. These styles aim to improve clarity, structure, and informational density while preserving the original semantic content. The resulting synthetic corpora are then used for continued pretraining of two English-centric base models: a 1.1B-parameter model derived from TinyLLaMA, originally trained on approximately 1T English tokens, and a 7B-parameter model derived from LLaMA-2, originally trained on approximately 2T English tokens.

We evaluate all models on PoETa v2, a comprehensive Portuguese benchmark comprising 44 tasks spanning multiple linguistic and reasoning categories. This evaluation enables a fine-grained analysis of how rewriting and source data quality affect downstream performance across model scales.

Our results show that synthetic rewriting can significantly amplify the benefits of limited Portuguese pretraining, but that these gains are strongly modulated by the quality of the original data and by model scale. At the 7B scale, rewriting high-quality data yields a substantial gain over using the same data unmodified, while rewriting low-quality data provides only marginal improvements. At the 1.1B scale, the interaction is weaker and less consistent, suggesting that the amplification effect is scale-dependent. These findings highlight the importance of combining principled data selection with synthetic data generation, and suggest that rewriting is most effective when applied as a multiplier of already good data rather than as a replacement for quality filtering.
\section{Related Work}

\subsection{Linguistic and cultural disparities in LLMs}

Recent benchmarks have documented systematic disparities in LLM performance across languages and regions, motivating targeted data interventions for underrepresented languages. BLEnD~\cite{lee2024blend} measures models' understanding of everyday activities across countries, showing that performance correlates with digital representation. WorldBench~\cite{moayeri2024worldbench} evaluates factual recall using socioeconomic data and finds higher error rates for underrepresented regions. TiEBe~\cite{almeida2025tiebe} observes a strong correlation between a country's GDP and model recall of notable events. INCLUDE~\cite{romanou2024include} constructs a large-scale evaluation suite of nearly 200,000 QA pairs from local exam sources across 44 languages, demonstrating that translating English benchmarks is insufficient because it ignores the cultural and regional knowledge specific to each language. ALM-Bench~\cite{vayani2025all} extends this analysis to the multimodal setting, evaluating models across 100 languages and reporting accuracy drops from 88.4\% in English to 50.8\% in low-resource languages for GPT-4o. Together, these findings establish that performance gaps are pervasive and not addressed by simply scaling English-centric models. For languages like Portuguese, where large-scale monolingual corpora remain comparatively scarce, continued pretraining on limited target-language data is often the most practical path to narrowing these gaps, raising the question of whether synthetic data transformations can further amplify the benefits of such constrained data budgets.

\subsection{Continued pretraining for Portuguese}
The development of Portuguese language models has evolved from masked language models such as BERTimbau \cite{bertimbau} and PTT5 \cite{ptt5v2} toward large-scale generative architectures. A common and cost-effective strategy is to take an English-dominant base model and continue pretraining it on Portuguese data. Sabiá \cite{sabia} extends LLaMA 7B and 65B with approximately 10B Portuguese tokens and observes substantial downstream gains despite the modest data volume. Cabrita \cite{cabrita} follows a similar recipe with OpenLLaMA 3B but additionally adapts the tokenizer embeddings to better represent Portuguese, training on roughly 7B tokens from mC4-PT. Curió-Edu 7B \cite{almeida2025curio} investigates the role of data selection in continued pretraining, showing that a LLaMA-2 7B model trained on just 10B tokens filtered for educational and STEM content can outperform a variant trained on 100B unfiltered tokens from the same corpus — highlighting that quality-based curation can be more effective than scale alone. More recently, Tucano~2 \cite{correa2026tucano2} performs continued pretraining of Qwen3 models for Portuguese, combining a revised corpus with synthetic augmentation data and post-training stages covering instruction following, tool use, and chain-of-thought reasoning.

Our work builds directly on the findings of Curió-Edu 7B, which established that data quality is a decisive factor in continued pretraining for Portuguese. Here, we ask the natural follow-up question: given that curated data is more effective, can synthetic rewriting further amplify its benefits — and does the quality of the original data matter when rewriting is applied? Using ClassiCC-PT \cite{almeida2025classicc} as the data source and two base models at different scales — TinyLLaMA 1.1B and LLaMA-2 7B — we construct controlled experimental conditions that isolate the interaction between source quality and rewriting, a dimension that, to our knowledge, has not been systematically investigated for Portuguese.

\subsection{Synthetic rewriting and transformation for pretraining}

In parallel, several recent works explore data transformation as a mechanism to improve pretraining efficiency and downstream performance without relying exclusively on more raw data. One prominent approach is style-conditioned rewriting of web documents. WRAP~\cite{maini2024rephrasing} proposes ``Web Rephrase Augmented Pre-training,'' which generates paraphrases of web text in styles such as Wikipedia-like and Q\&A-like formats and mixes them with real text; it reports improved data and computational efficiency when applied to noisy corpora such as C4. More recently, REWIRE~\cite{nguyen2025recycling} focuses on transforming low-quality documents that would otherwise be discarded by filtering pipelines, producing rewritten text that can be mixed with filtered high-quality data to yield consistent downstream improvements. Importantly for our study design, REWIRE also analyzes how rewrite quality relates to the original data, suggesting that rewritten text can remain useful even when sourced from lower-quality inputs.

Other transformation pipelines target specialized domains. Nemotron-CC~\cite{su2025nemotron} combines classifier ensembling with synthetic rephrasing to improve trade-offs between quality and quantity for long-horizon pretraining, showing that synthetic transformations can complement filtering rather than merely replacing it. In math and code, ``transform-and-retain'' rewriting pipelines such as SwallowCode and SwallowMath~\cite{fujii2025rewriting} systematically restructure, clean, and rewrite data to yield large gains in downstream reasoning and synthesis tasks under fixed token budgets, underscoring that rewriting can act as a strong multiplicative factor on data utility.

Despite this progress, most rewriting-based pretraining studies are conducted primarily in English and are often evaluated on English-centric benchmarks. In contrast, our work studies Portuguese continued pretraining under a fixed 10B-token budget and explicitly tests how gains from rewriting depend on the quality of the original source data, leveraging ClassiCC-PT’s STEM and Educational scoring~\cite{almeida2025classicc} to construct controlled subsets. This setup allows us to disentangle (i) the effect of original data quality, (ii) the effect of style-conditioned rewriting, and (iii) their interaction across two model scales, evaluated on PoETa~V2~\cite{almeida2025poeta}, a Portuguese-focused benchmark.

\section{Methodology}
\label{sec:methodology}

\subsection{Data source and subset construction}
Our experiments are based on \textbf{ClassiCC-PT}~\cite{almeida2025classicc}, a large-scale Portuguese corpus comprising approximately 120B tokens, annotated with continuous \textit{STEM} and \textit{Educational} quality scores ranging from 0 to 5. These scores are produced by classifiers trained on 120K documents annotated with GPT-4o. The annotations enable controlled data selection based on document-level quality estimates.

From this corpus, we construct two subsets, each containing \textbf{10B tokens measured prior to rewriting}:
\begin{itemize}
    \item \textbf{High-quality subset}, composed of documents with either STEM or Educational scores greater than 2.5;
    \item \textbf{Low-quality subset}, composed of documents with scores between 0.5 and 2.0;
\end{itemize}


\subsection{Synthetic rewriting}
For each of the two subsets, we generate synthetic data via large-scale document rewriting. We use the instruction-tuned \textbf{Qwen-2.5-7B} model to rewrite each document into four distinct styles inspired by prior work on web text rephrasing~\cite{maini2024rephrasing}. The choice of a 7B rewriting model was motivated by computational budget constraints and by the goal of studying whether a relatively small model is sufficient to produce beneficial rewrites---an important practical consideration for large-scale data pipelines. The four styles are:
\begin{itemize}
    \item \textbf{Easy}: simplifies the original text to a lower reading level, using shorter sentences and more common vocabulary while preserving the core information;
    \item \textbf{Medium}: rewrites the text in a clean, encyclopedic style similar to Wikipedia, improving clarity and structure without altering the difficulty level;
    \item \textbf{Hard}: produces a more technical and formal version of the original, using domain-specific terminology and complex sentence structures;
    \item \textbf{QA}: transforms the document into a question-and-answer format, extracting key points as questions with corresponding answers.
\end{itemize}

The rewriting process is \textbf{stochastic}, introducing lexical and structural variation across rewritten outputs. No explicit length control is imposed; consequently, rewritten documents may differ substantially in length from their originals. Combining all four styles, each 10B-token subset yields approximately \textbf{30B tokens of synthetic data}. The synthetic rewrites are then concatenated with the \textbf{10B tokens of original} (unrewritten) documents, producing approximately \textbf{30B tokens per condition}. Token counts after rewriting vary slightly across subsets and styles.

\subsection{Base models}
We study two English-centric pretrained models at different scales:
\begin{itemize}
    \item A \textbf{1.1B-parameter model} based on TinyLLaMA~\cite{zhang2024tinyllama}, pretrained on approximately 1T English tokens;
    \item A \textbf{7B-parameter model} based on LLaMA-2~\cite{touvron2023llama2}, pretrained on approximately 2T English tokens.
\end{itemize}

For both models, we retain the original tokenizer and vocabulary, and perform no tokenizer adaptation or retraining.

\subsection{Continued pretraining}
Each model is continued-pretrained separately on each data condition. For conditions \textit{with} rewriting, the training data consists of approximately 30B synthetic tokens (from all four rewrite styles) combined with the 10B original tokens, totaling approximately \textbf{40B unique tokens} per condition. For conditions \textit{without} rewriting, the same 10B original tokens are repeated for multiple epochs to match the 40B token budget. This design ensures a consistent compute budget across all conditions, but the rewritten conditions benefit from substantially more unique content---an important consideration when interpreting the results.

Following the training configuration of~\cite{almeida2025classicc}, all runs use the AdamW optimizer with a learning rate of $3 \times 10^{-4}$, a cosine decay schedule with 5\% warmup, and a sequence length of 4{,}096 tokens. All models are trained using the same training configuration within each model scale, varying only the underlying data partition. This design isolates the effects of (i) source data quality and (ii) synthetic rewriting, while controlling for training budget and model architecture.

\subsection{Evaluation}
We evaluate all models using \textbf{PoETa v2}~\cite{almeida2025poeta}, a comprehensive Portuguese benchmark comprising \textbf{44 tasks} spanning multiple linguistic and reasoning domains. All results are reported using the \textbf{Normalized Performance Metric (NPM)}, which rescales each task's score to the $[0, 100]$ range relative to a random baseline (NPM~$= 0$) and a perfect score (NPM~$= 100$), enabling meaningful averaging across tasks with different native scales and chance levels. We report:
\begin{itemize}
    \item \textbf{Overall average performance} across all tasks;
    \item \textbf{Category-level performance}, following the PoETa v2 task groupings (e.g., text understanding, reasoning).
\end{itemize}

We do not perform statistical significance testing, and results are reported based on single runs per configuration. This evaluation protocol enables a fine-grained comparison of how rewriting and source data quality affect downstream performance across task types and model scales.

\section{Results}

We present results for both model scales, examining overall performance trajectories as well as category-level patterns. All scores are reported as Normalized Performance Metric (NPM) on PoETa~V2, averaged across all 44~tasks unless otherwise noted.

\subsection{7B model: overall performance}

\begin{figure}[!htb]
\centering
\includegraphics[width=1\linewidth]{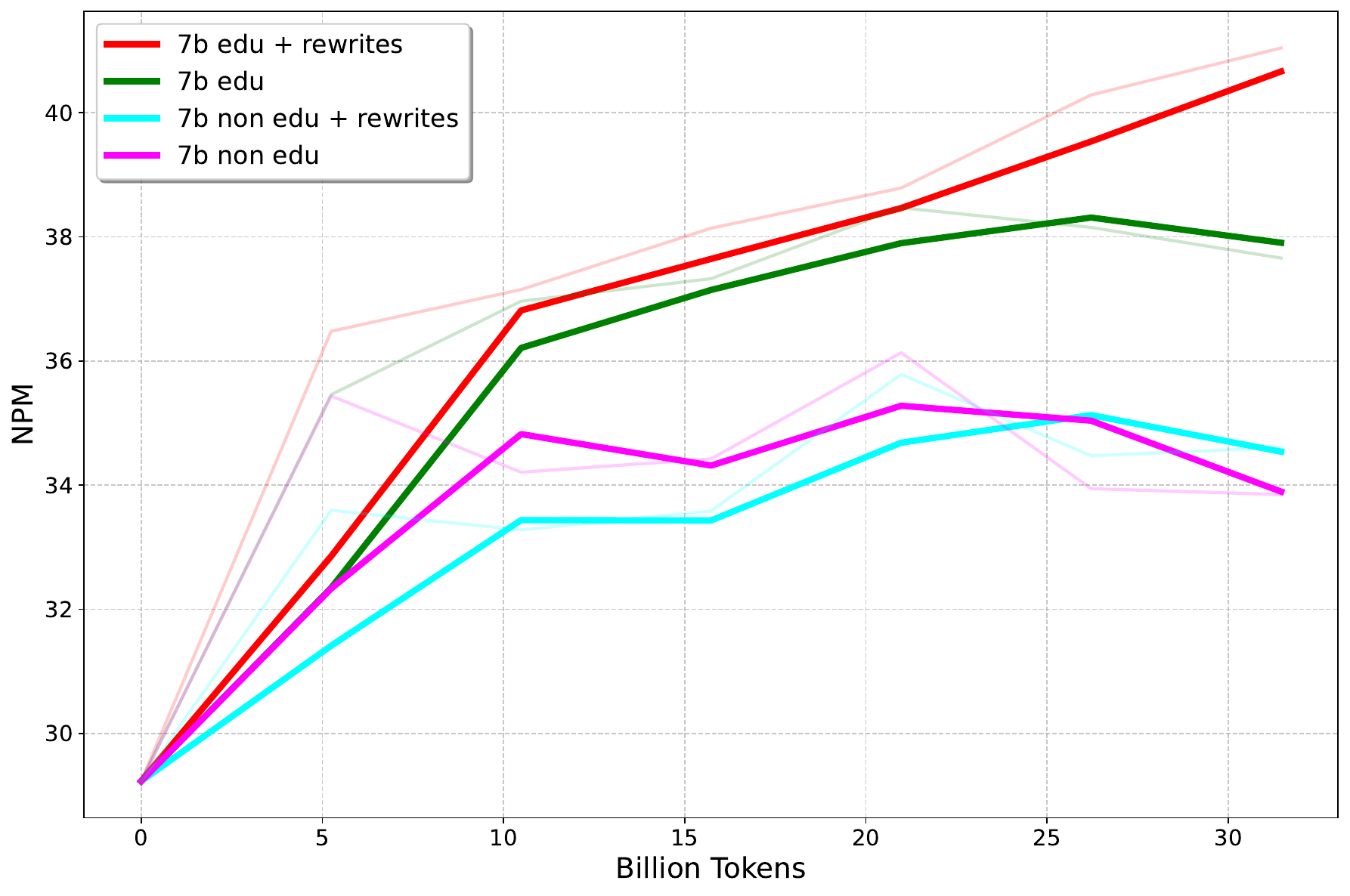}
\caption{Average NPM in PoETa~V2 for the 7B model across four experimental conditions as a function of training tokens.}
\label{fig:7b_main_results}
\end{figure}

Figure~\ref{fig:7b_main_results} shows the average NPM across all PoETa~V2 tasks as a function of training tokens for the 7B model under each experimental condition. A clear hierarchy emerges early in training and persists throughout. The high-quality subset with rewriting (\textit{edu + rewrites}) consistently achieves the highest performance, reaching 41.0~NPM after 30B tokens and still trending upward. The high-quality subset without rewriting (\textit{edu}) is the second-best condition, peaking at approximately 38.5~NPM around 25B tokens before showing signs of saturation. The two low-quality conditions---\textit{non-edu + rewrites} and \textit{non-edu}---cluster together at substantially lower performance levels, reaching approximately 35.8 and 35.2~NPM respectively.

The most notable pattern is the \textbf{asymmetric effect of rewriting across quality tiers}. On the high-quality subset, rewriting provides a gain of approximately +3.4~NPM over using the same data without rewriting (41.0 vs.\ 37.7 at 30B tokens). On the low-quality subset, rewriting yields a much smaller and less consistent gain of approximately +0.5~NPM. This asymmetry suggests that synthetic rewriting acts primarily as a \textit{multiplier} of existing data quality rather than as a mechanism for compensating poor source quality.

An important caveat is that the rewritten conditions contain substantially more \textit{unique} tokens than their non-rewritten counterparts: approximately 40B unique tokens (30B synthetic + 10B original) versus 10B unique tokens repeated four times. Part of the rewriting gain may therefore stem from increased data diversity rather than improved data quality per se. Nonetheless, if unique token count were the dominant factor, the gain should be comparable across quality tiers, whereas we observe a much larger gain for high-quality data. This suggests that the quality of the rewritten content---not merely its quantity---drives the differential improvement.

Additionally, the \textit{edu + rewrites} curve does not appear to have converged at 30B tokens, whereas the low-quality conditions plateau earlier (around 15--20B tokens). This suggests that high-quality rewritten data sustains learning over longer training horizons.

\subsection{7B model: category-level analysis}

\begin{figure*}[!htb]
\centering
\subfloat[Brazil]{\includegraphics[width=0.31\linewidth]{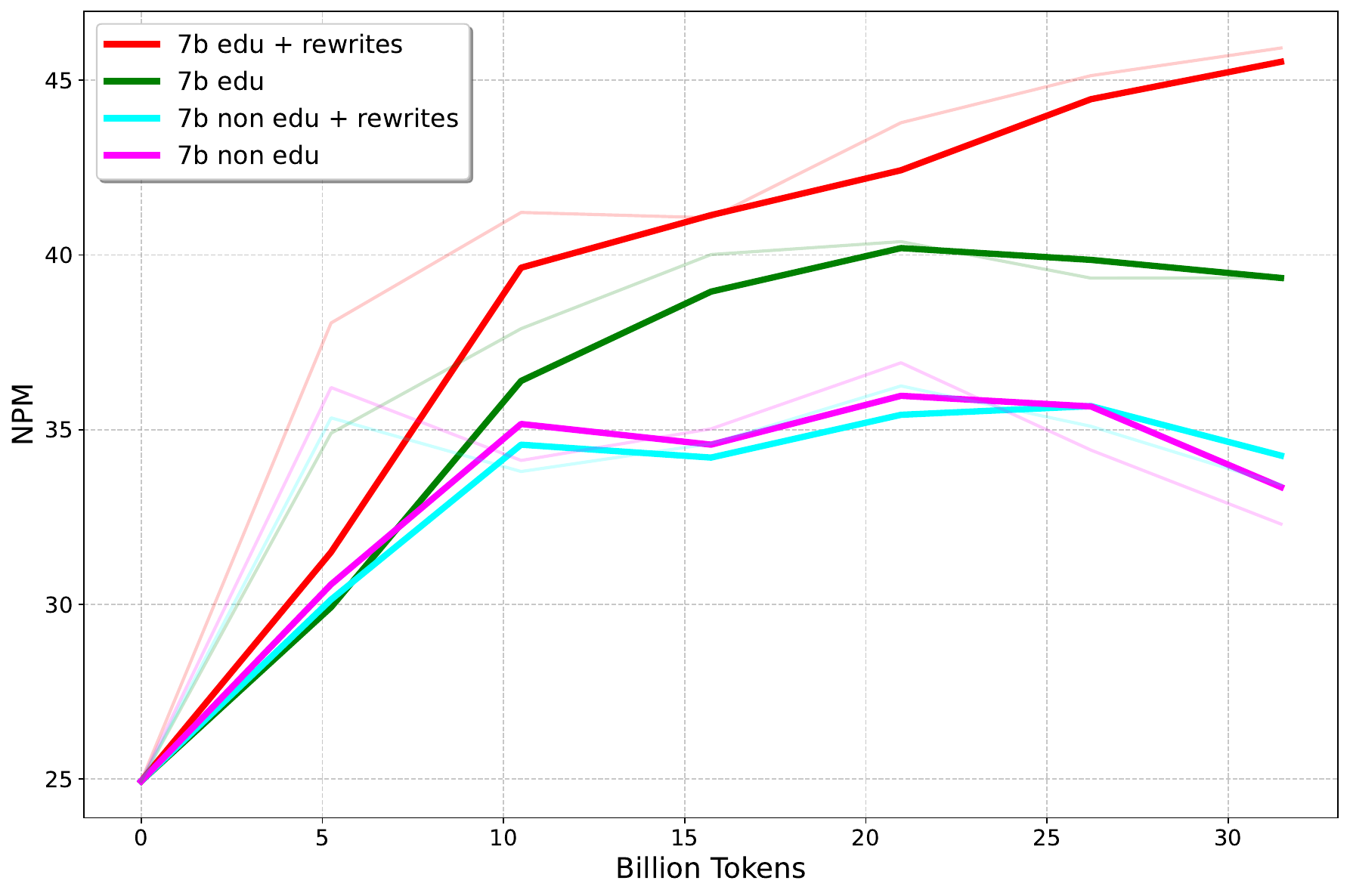}} \hspace*{0.5cm}
\subfloat[Text Understanding]{\includegraphics[width=0.31\linewidth]{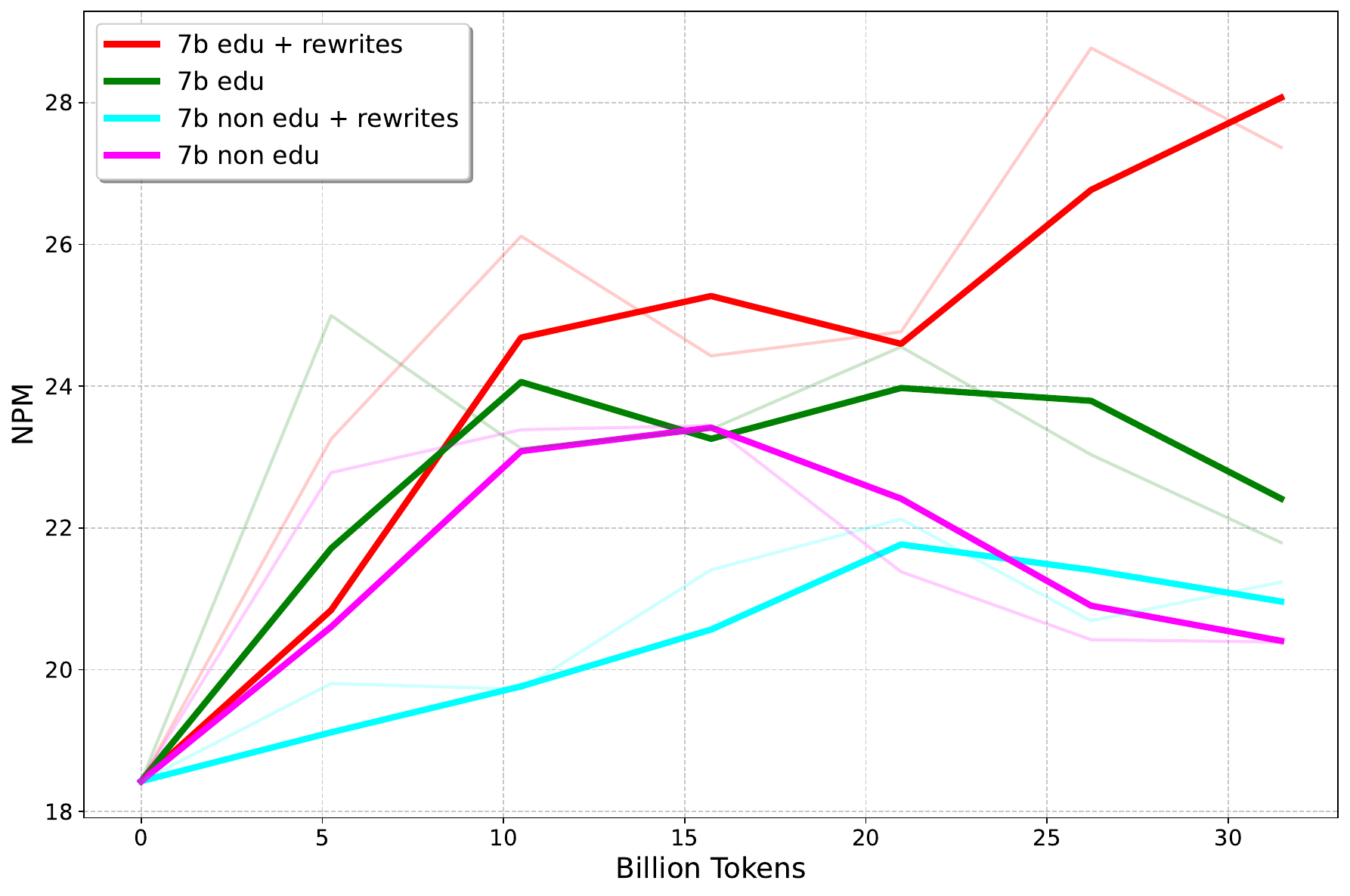}} \hspace*{0.5cm}
\subfloat[Exams]{\includegraphics[width=0.31\linewidth]{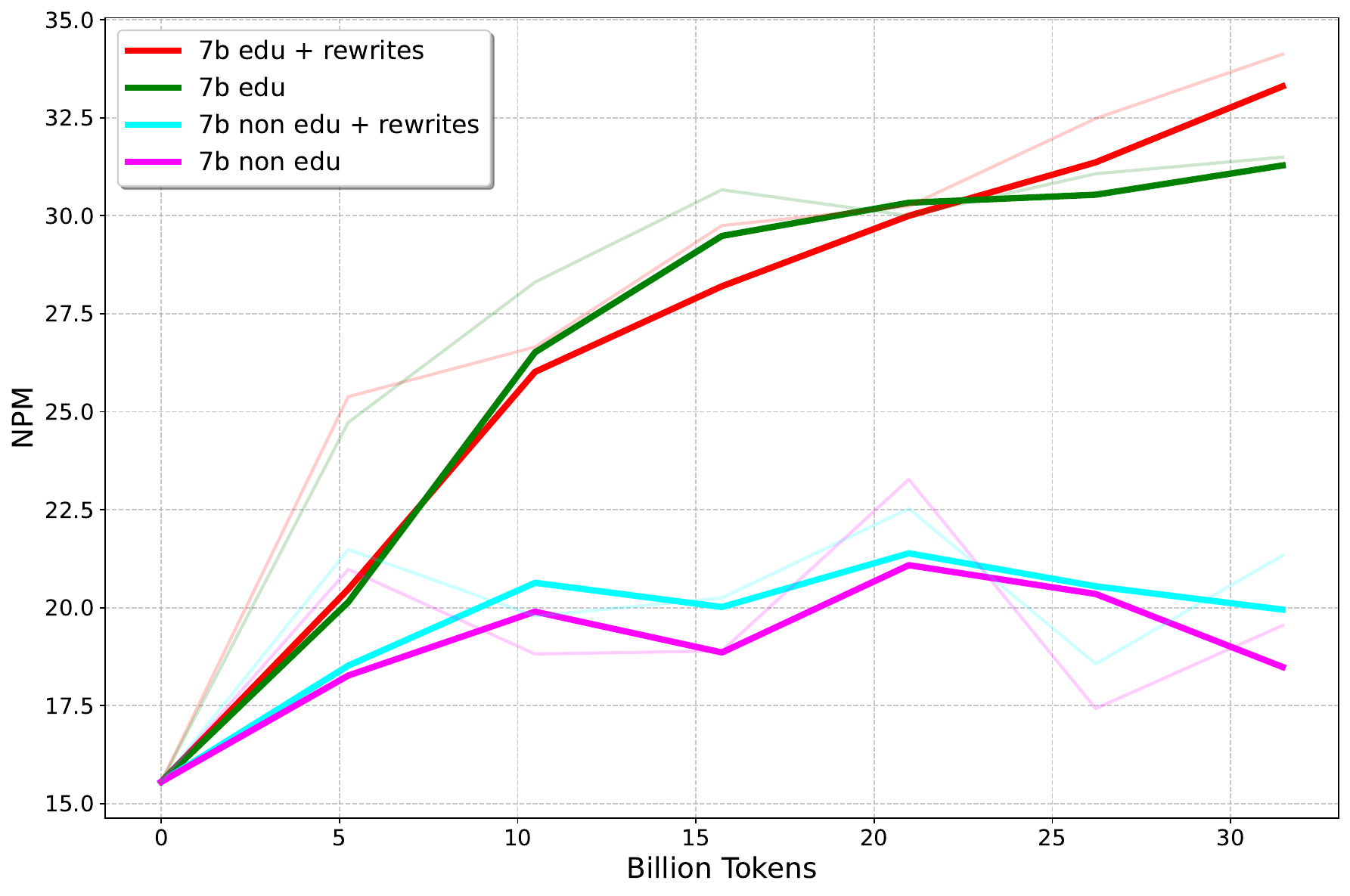}} \\
\subfloat[Reasoning]{\includegraphics[width=0.31\linewidth]{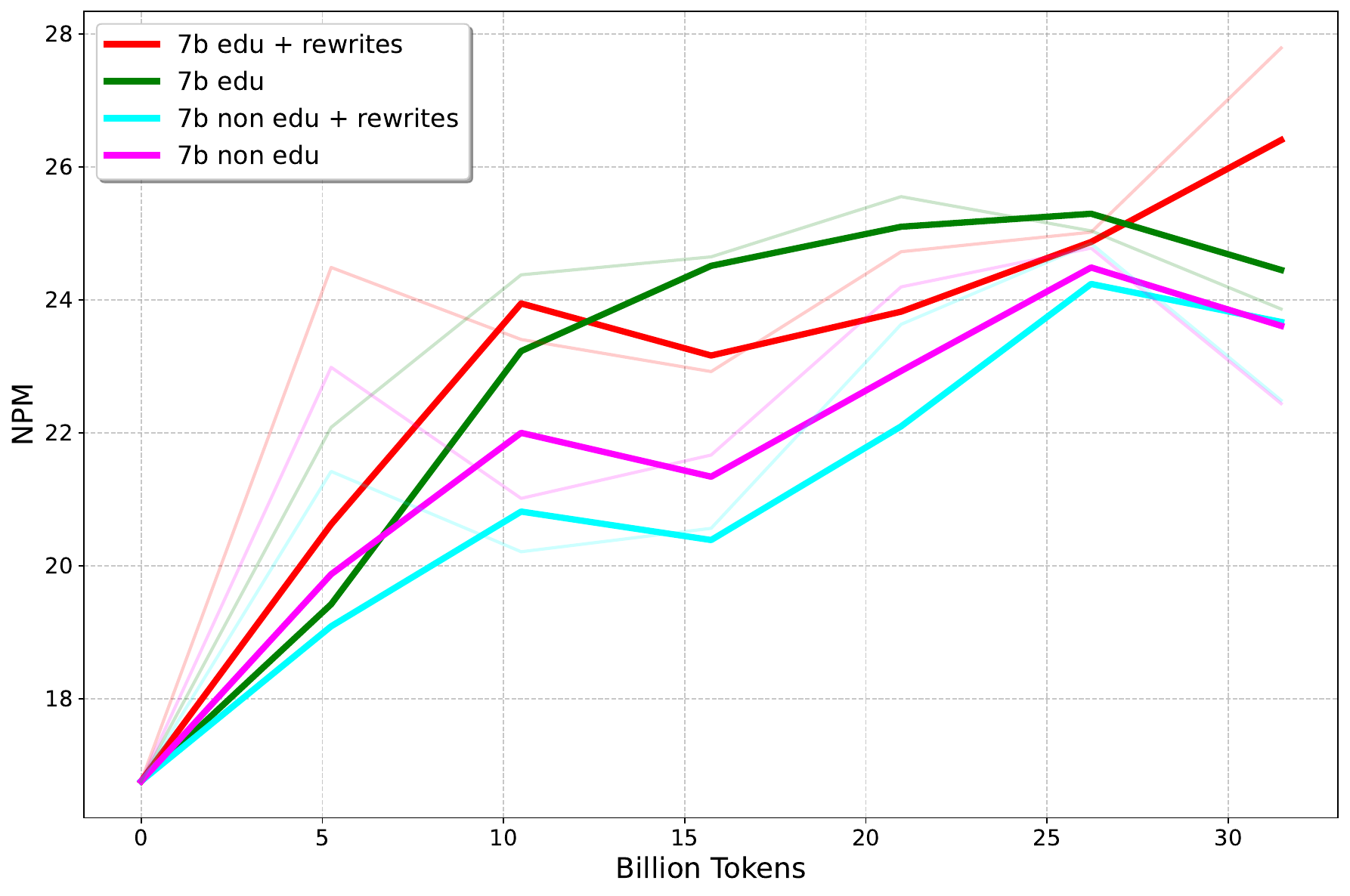}} \hspace*{0.5cm}
\subfloat[Common Sense]{\includegraphics[width=0.31\linewidth]{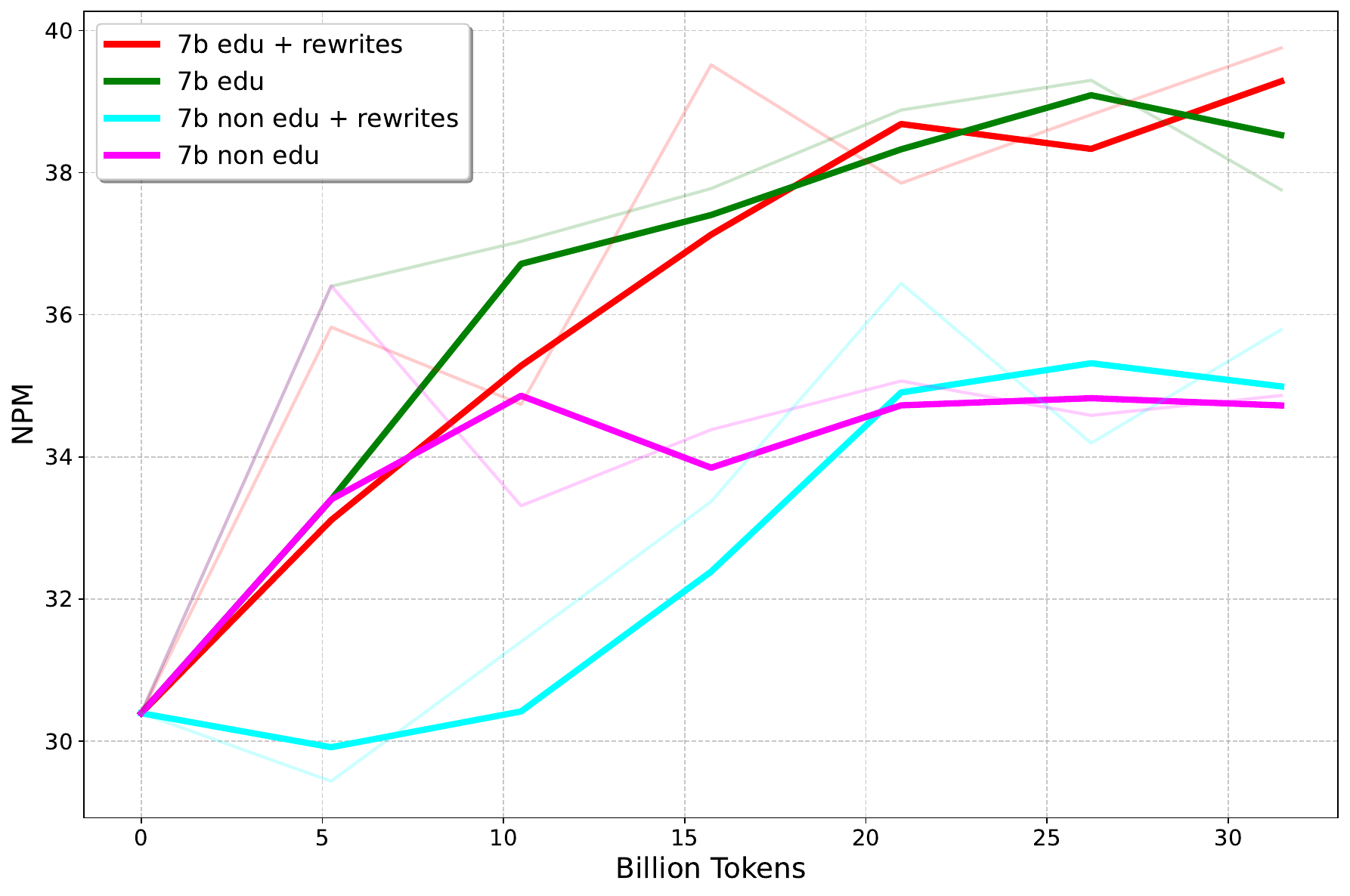}} \hspace*{0.5cm}
\subfloat[Math]{\includegraphics[width=0.31\linewidth]{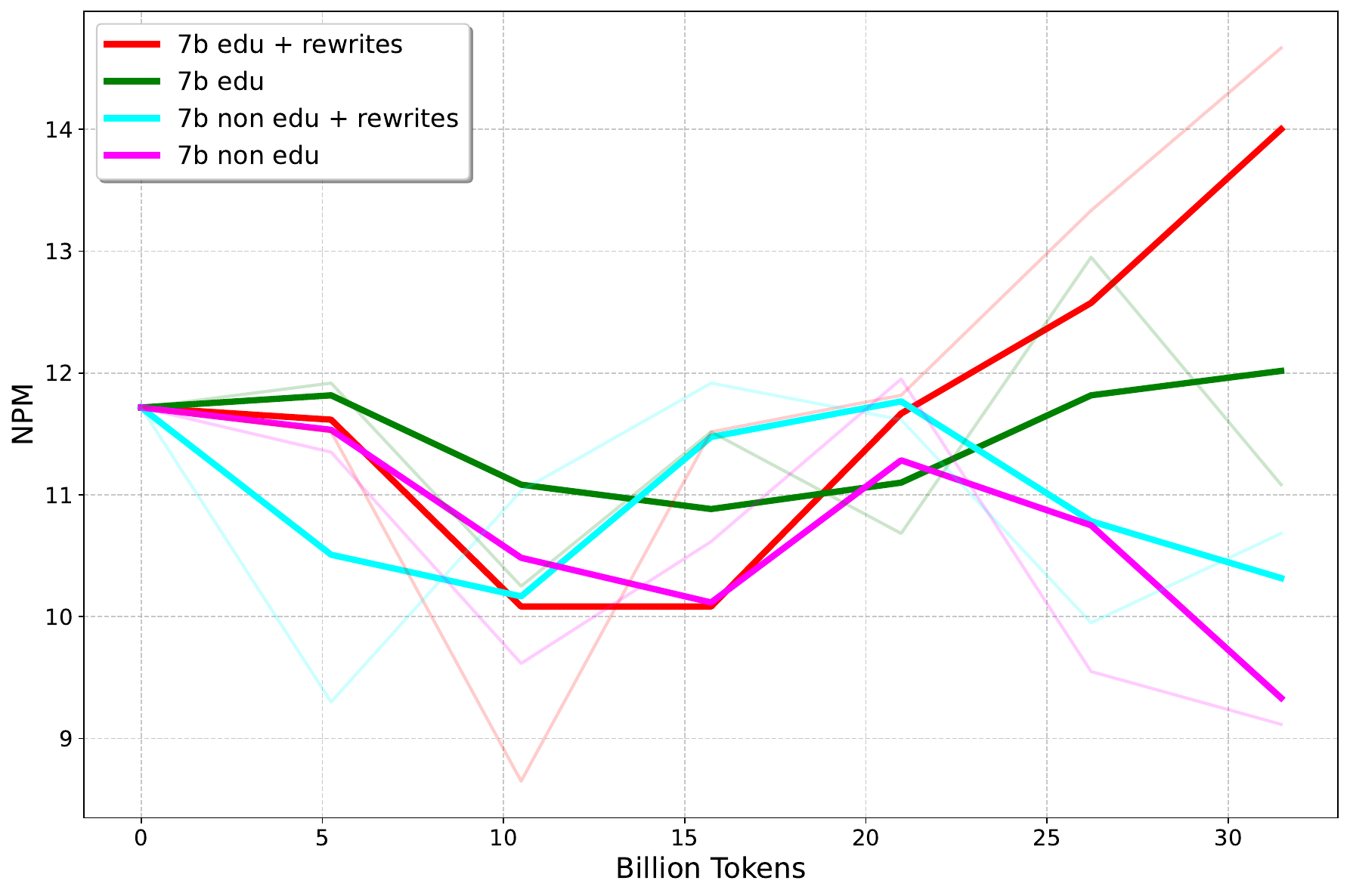}} \\
\subfloat[Social Media]{\includegraphics[width=0.31\linewidth]{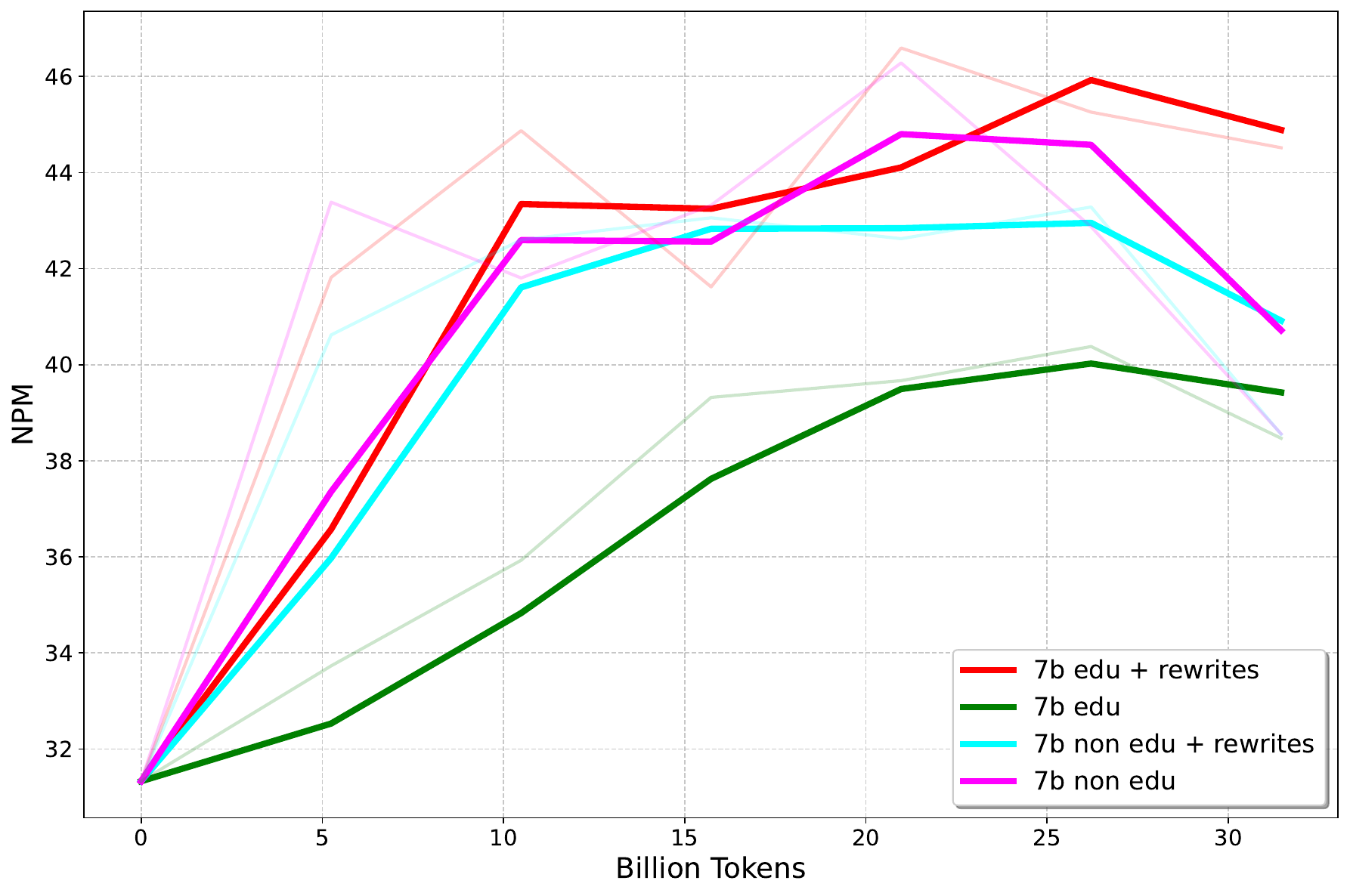}} \hspace*{0.5cm}
\subfloat[General Knowledge]{\includegraphics[width=0.31\linewidth]{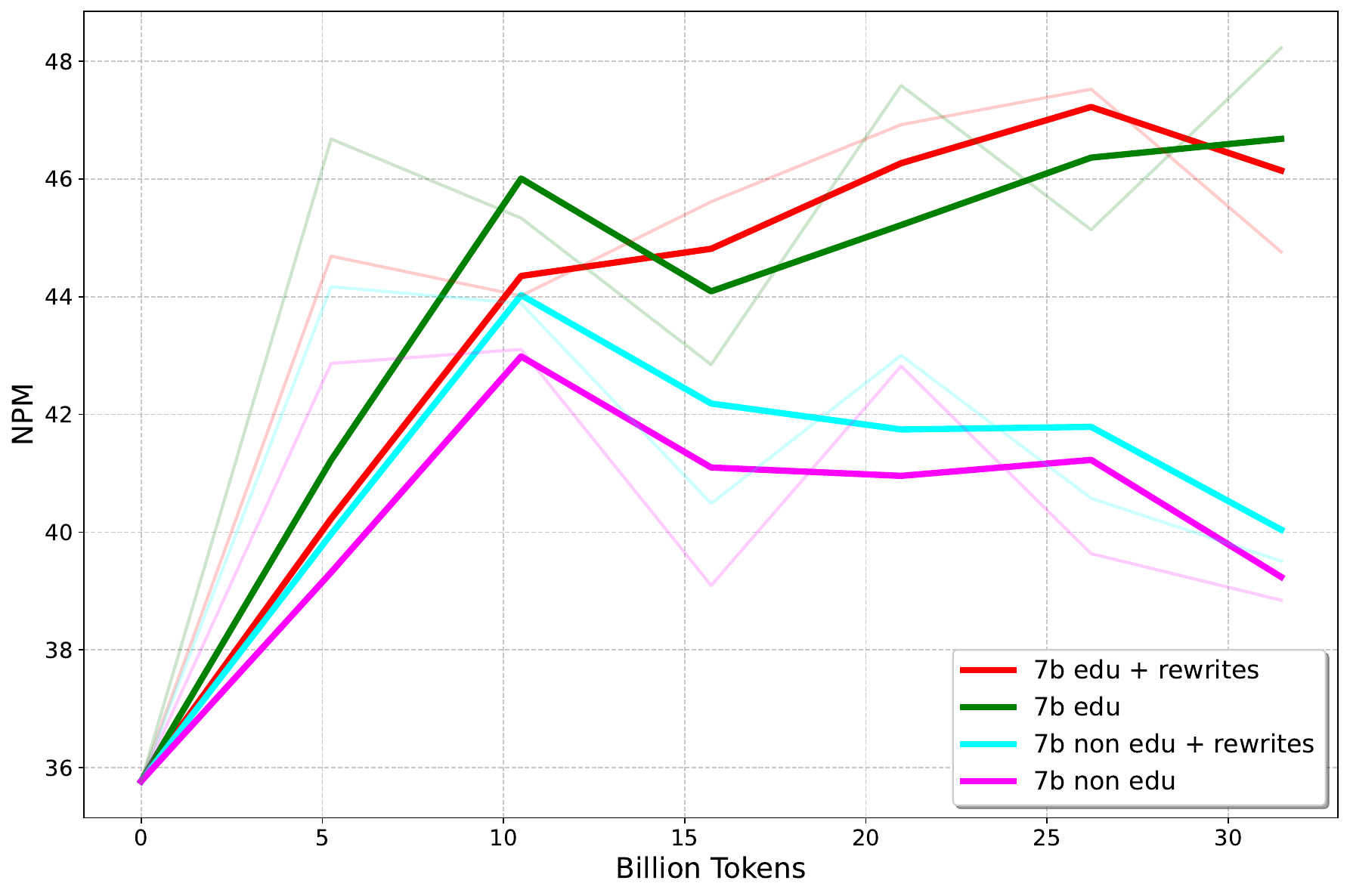}} \hspace*{0.5cm}
\subfloat[Ethics]{\includegraphics[width=0.31\linewidth]{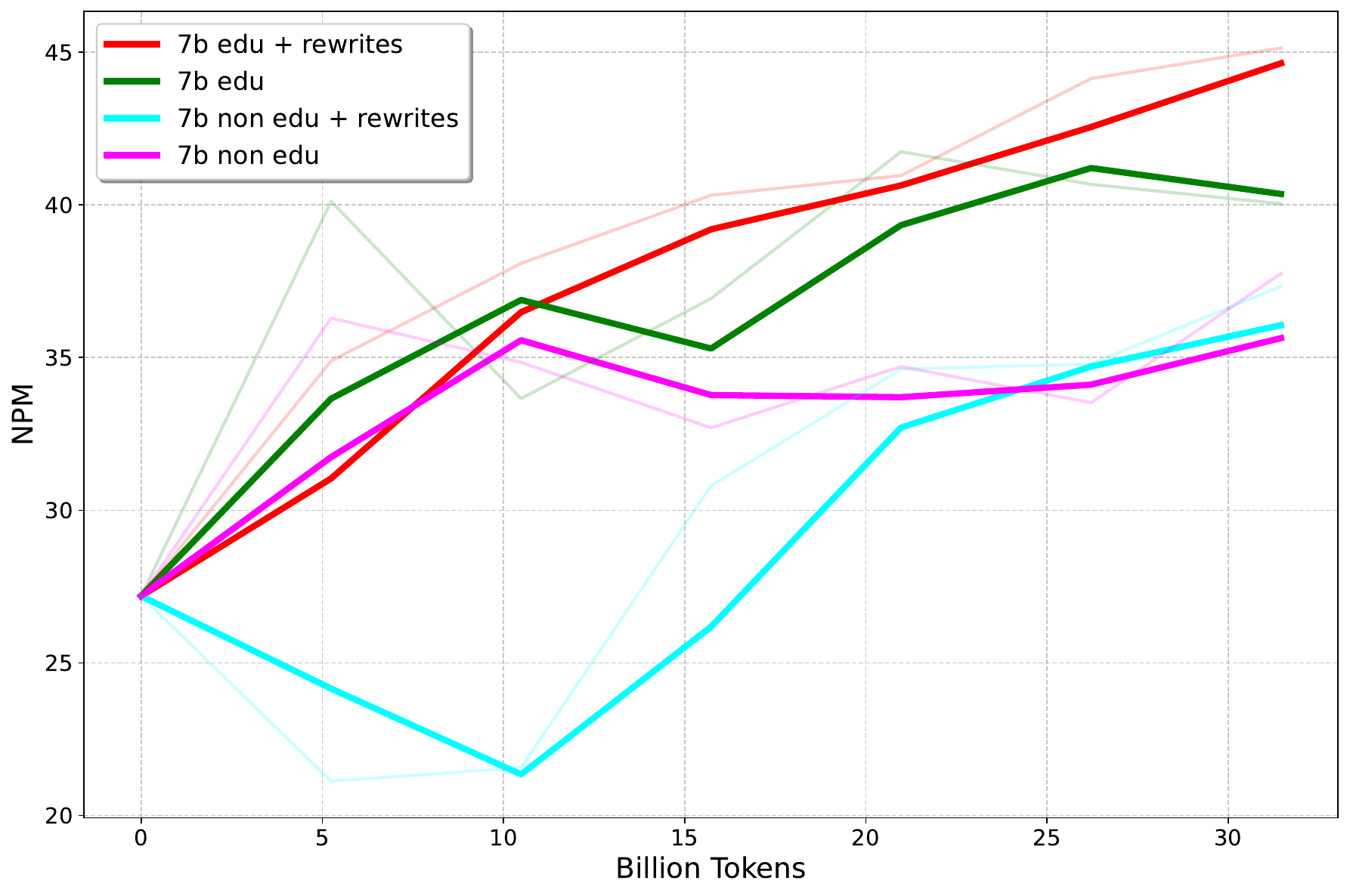}}
\caption{7B model: average NPM across task subcategories as a function of training tokens. The nine subcategories shown are those with the largest number of tasks in PoETa~V2.}
\label{fig:7b_subcategories}
\end{figure*}

Figure~\ref{fig:7b_subcategories} breaks down the 7B results by task subcategory. The quality-rewriting interaction varies substantially across task types.

\textbf{Exams and Brazil-specific tasks} show the largest quality effects. In the Exams subcategory, \textit{edu + rewrites} reaches 34.1~NPM versus 21.3~NPM for \textit{non-edu + rewrites}---a gap of nearly 13 points. Similarly, Brazil-specific tasks show \textit{edu + rewrites} at 45.9~NPM compared to 33.4~NPM for \textit{non-edu + rewrites}. These categories involve structured knowledge and culturally grounded content, where the original data quality appears most consequential.

\textbf{Ethics} is a category where rewriting provides a substantial boost regardless of source quality: \textit{edu + rewrites} reaches 45.1~NPM, \textit{edu} alone reaches 40.0~NPM, and even \textit{non-edu + rewrites} (37.3~NPM) substantially outperforms \textit{non-edu} without rewriting (31.6~NPM).

\textbf{General Knowledge} presents a contrasting pattern: \textit{edu} without rewriting (48.2~NPM) actually outperforms \textit{edu + rewrites} (44.8~NPM), suggesting that rewriting may occasionally distort factual content or introduce noise for knowledge-intensive tasks.

\textbf{Social Media} tasks show high performance across all conditions, with \textit{non-edu} reaching 46.3~NPM---the highest among all conditions for this category---likely reflecting that lower-quality web data naturally contains social-media-style text.

\subsection{1.1B model: overall performance}

\begin{figure}[!htb]
\centering
\includegraphics[width=1\linewidth]{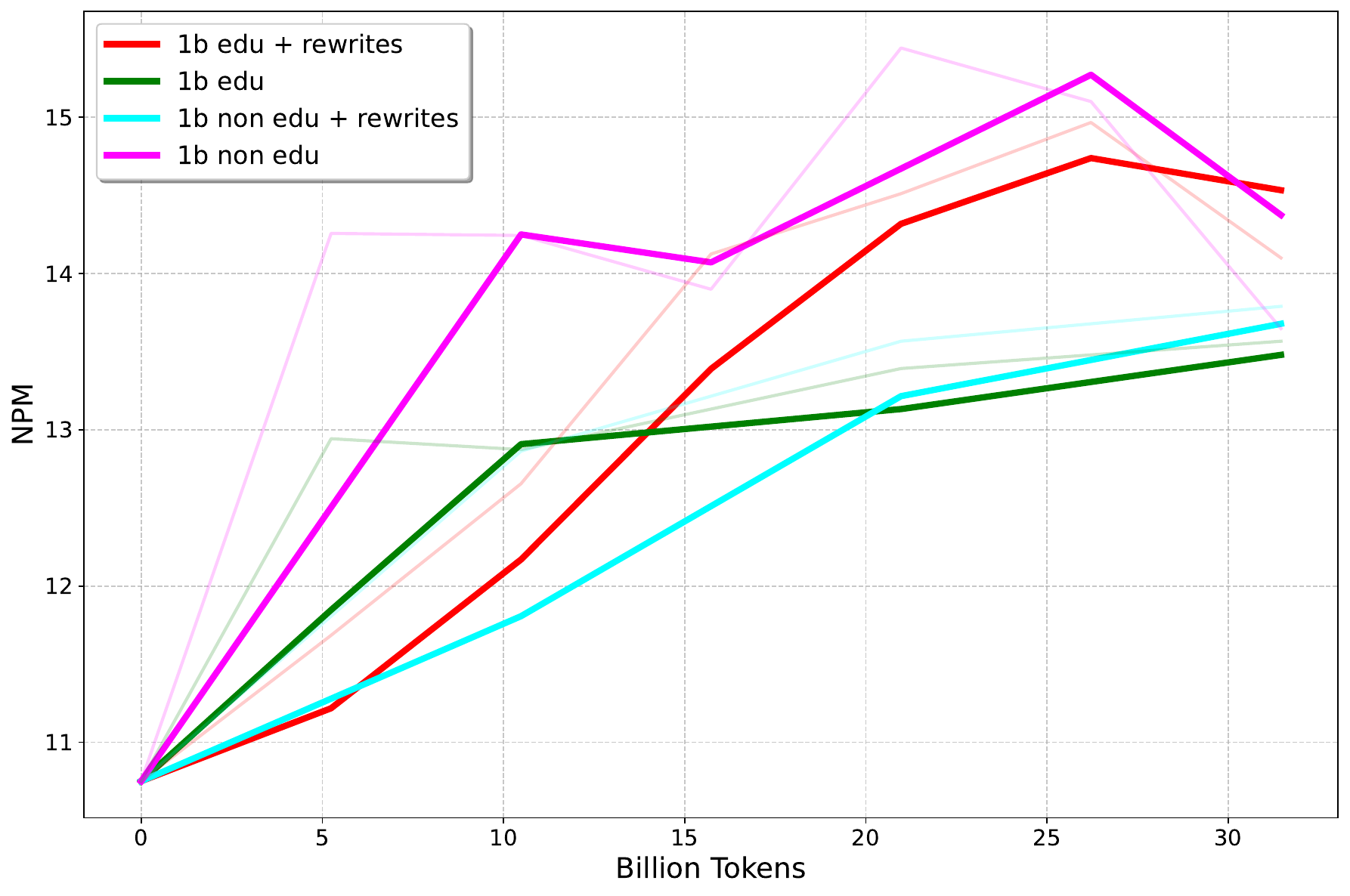}
\caption{Average NPM in PoETa~V2 for the 1.1B model across four experimental conditions as a function of training tokens.}
\label{fig:1b_main_results}
\end{figure}

Figure~\ref{fig:1b_main_results} shows the corresponding results for the 1.1B model. The overall performance landscape is substantially different from the 7B case: absolute NPM values are lower (ranging from approximately 11 to 15), and the separation between conditions is less pronounced and less consistent.

The \textit{edu + rewrites} condition peaks at approximately 15.0~NPM around 25B tokens, while the \textit{non-edu + rewrites} condition reaches approximately 13.8~NPM at 30B tokens. Notably, the \textit{non-edu} condition (without rewriting) performs comparably to \textit{edu + rewrites}, reaching 15.1~NPM at 20B tokens and exceeding the rewritten high-quality condition at several intermediate checkpoints. The curves cross multiple times, suggesting that at this smaller model scale, the quality-rewriting interaction observed at 7B does not reliably hold.


Several factors may explain this discrepancy. First, a 1.1B-parameter model may lack sufficient capacity to fully leverage the structured transformations introduced by a 7B rewriting model, effectively saturating on the added complexity. Second, the non-rewritten conditions train on the same 10B tokens repeated multiple times, which at this smaller scale may act as a form of regularization that benefits generalization. Third, the raw diversity present in unfiltered web data may provide a stronger learning signal at smaller scales, where memorizing diverse surface patterns is more beneficial than learning from cleaner but more homogeneous rewrites.

\subsection{1.1B model: category-level analysis}

\begin{figure*}[!htb]
\centering
\subfloat[Brazil]{\includegraphics[width=0.31\linewidth]{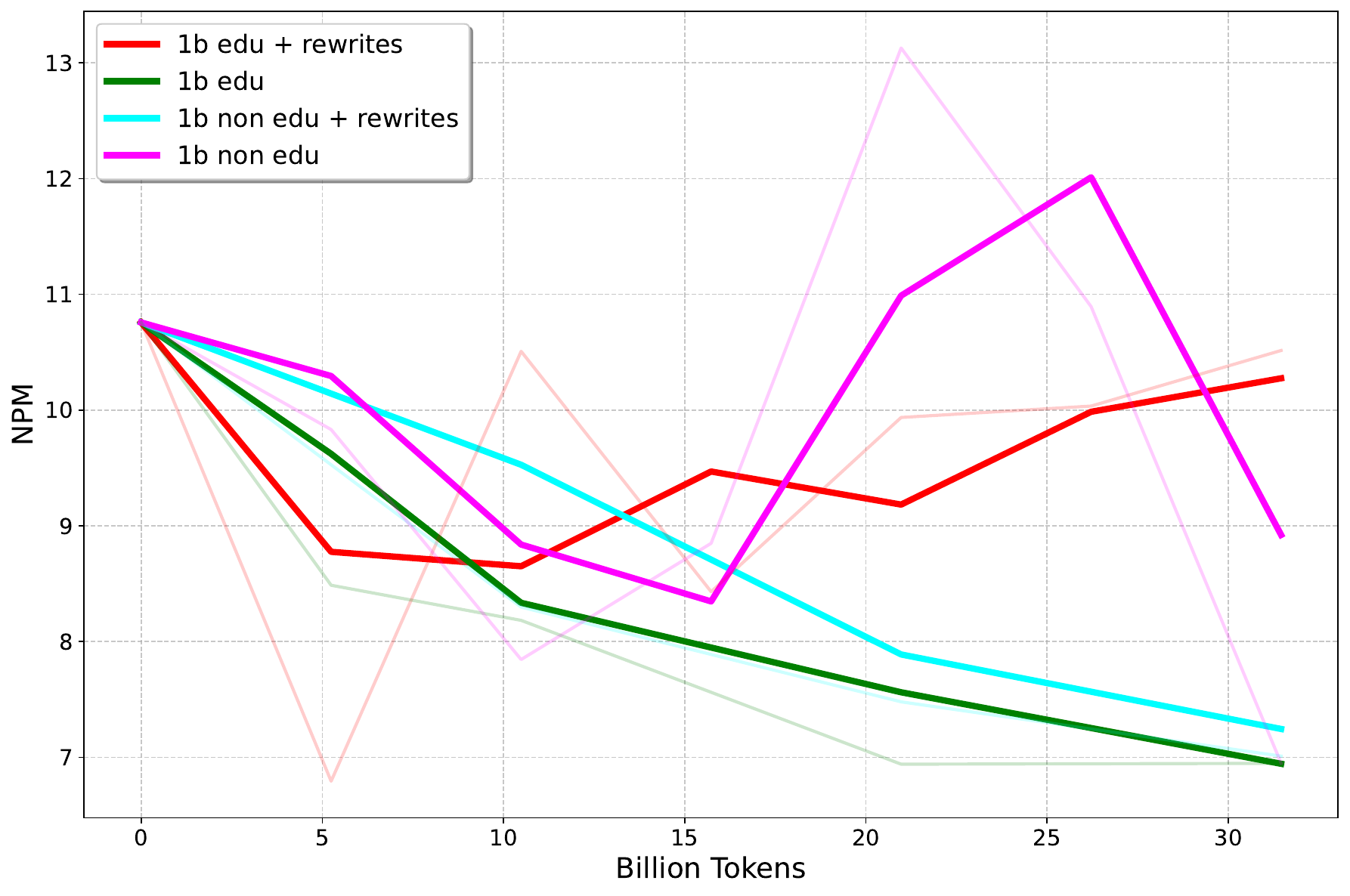}} \hspace*{0.5cm}
\subfloat[Text Understanding]{\includegraphics[width=0.31\linewidth]{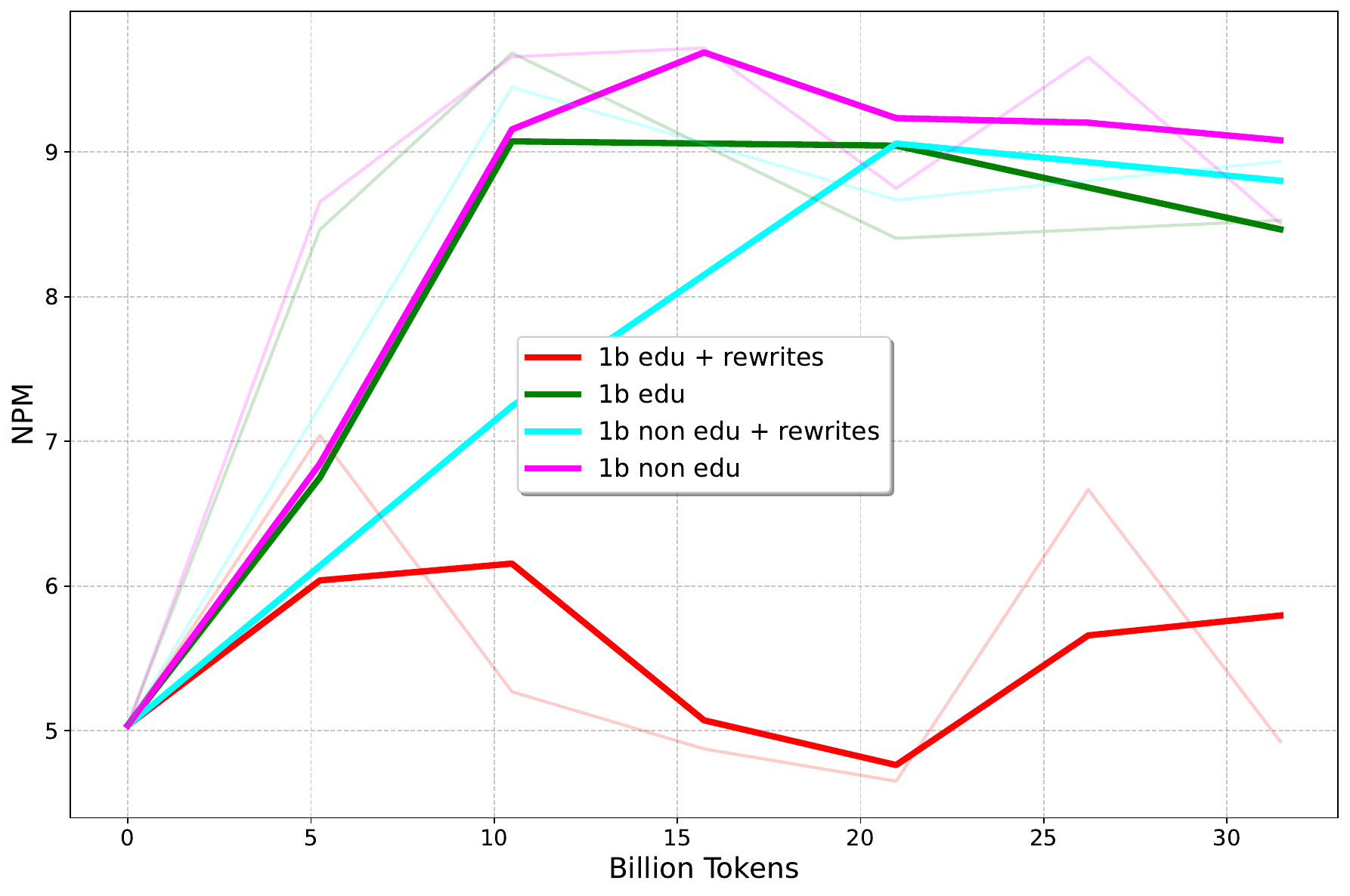}} \hspace*{0.5cm}
\subfloat[Exams]{\includegraphics[width=0.31\linewidth]{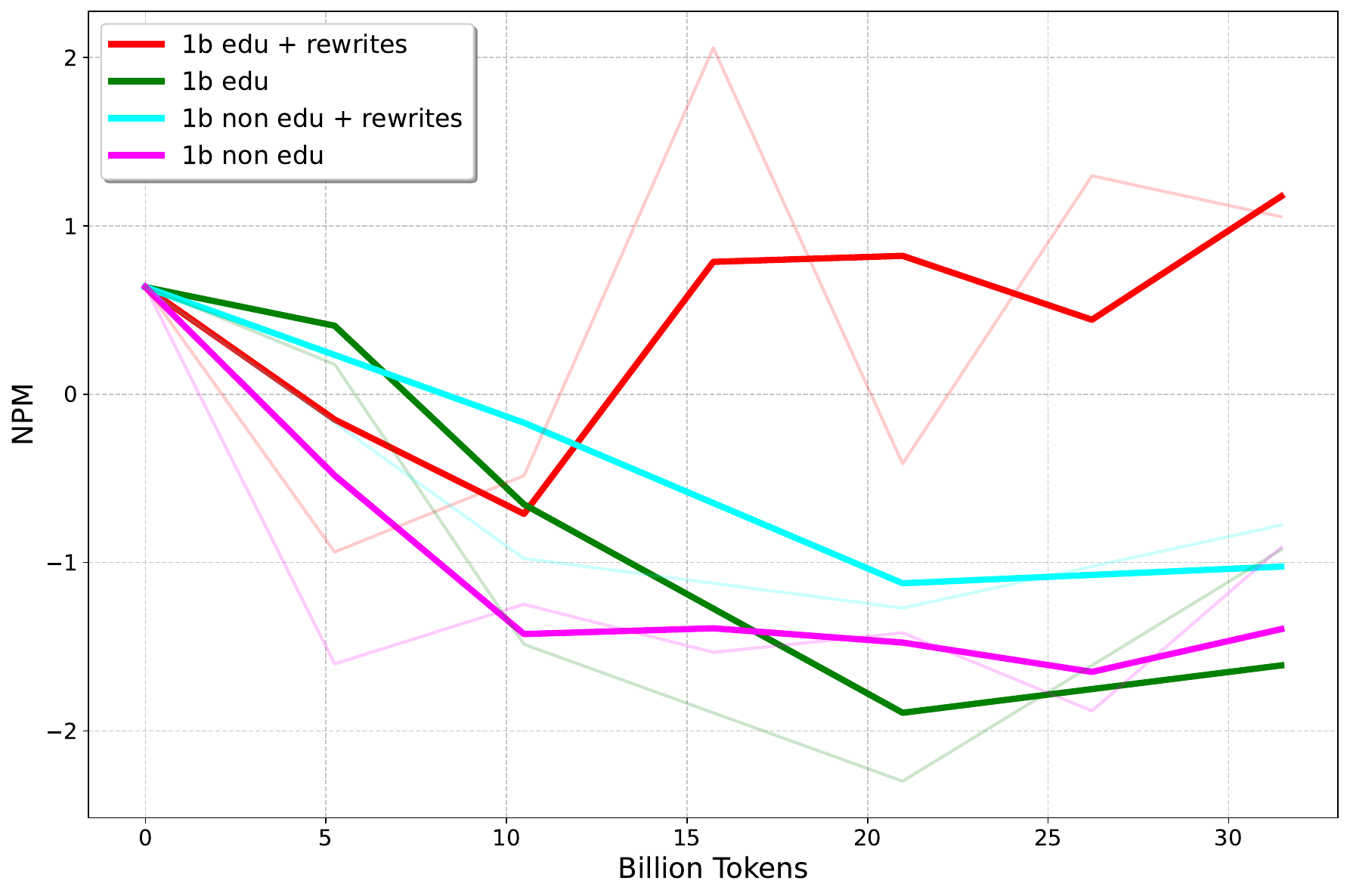}} \\
\subfloat[Reasoning]{\includegraphics[width=0.31\linewidth]{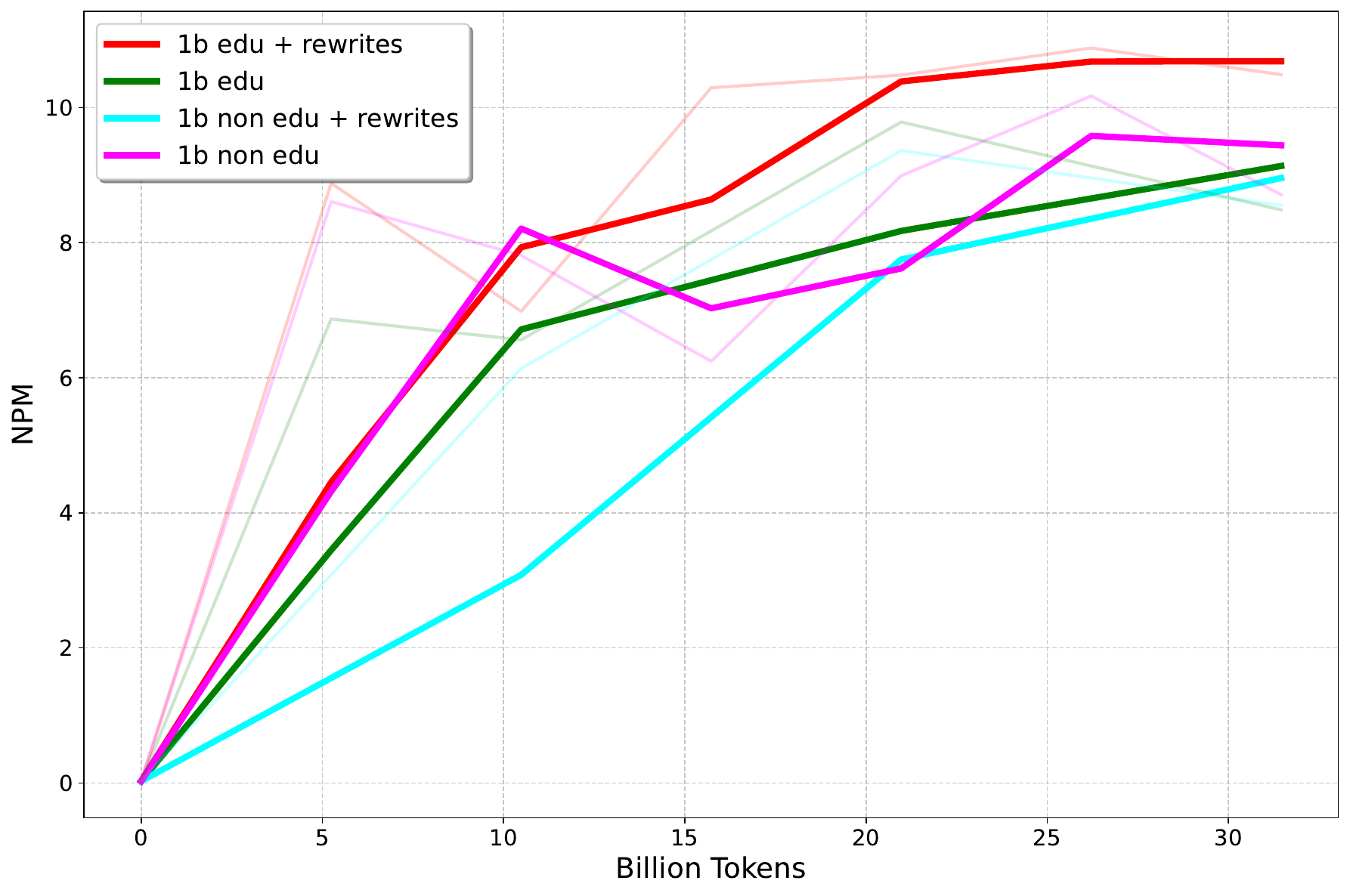}} \hspace*{0.5cm}
\subfloat[Common Sense]{\includegraphics[width=0.31\linewidth]{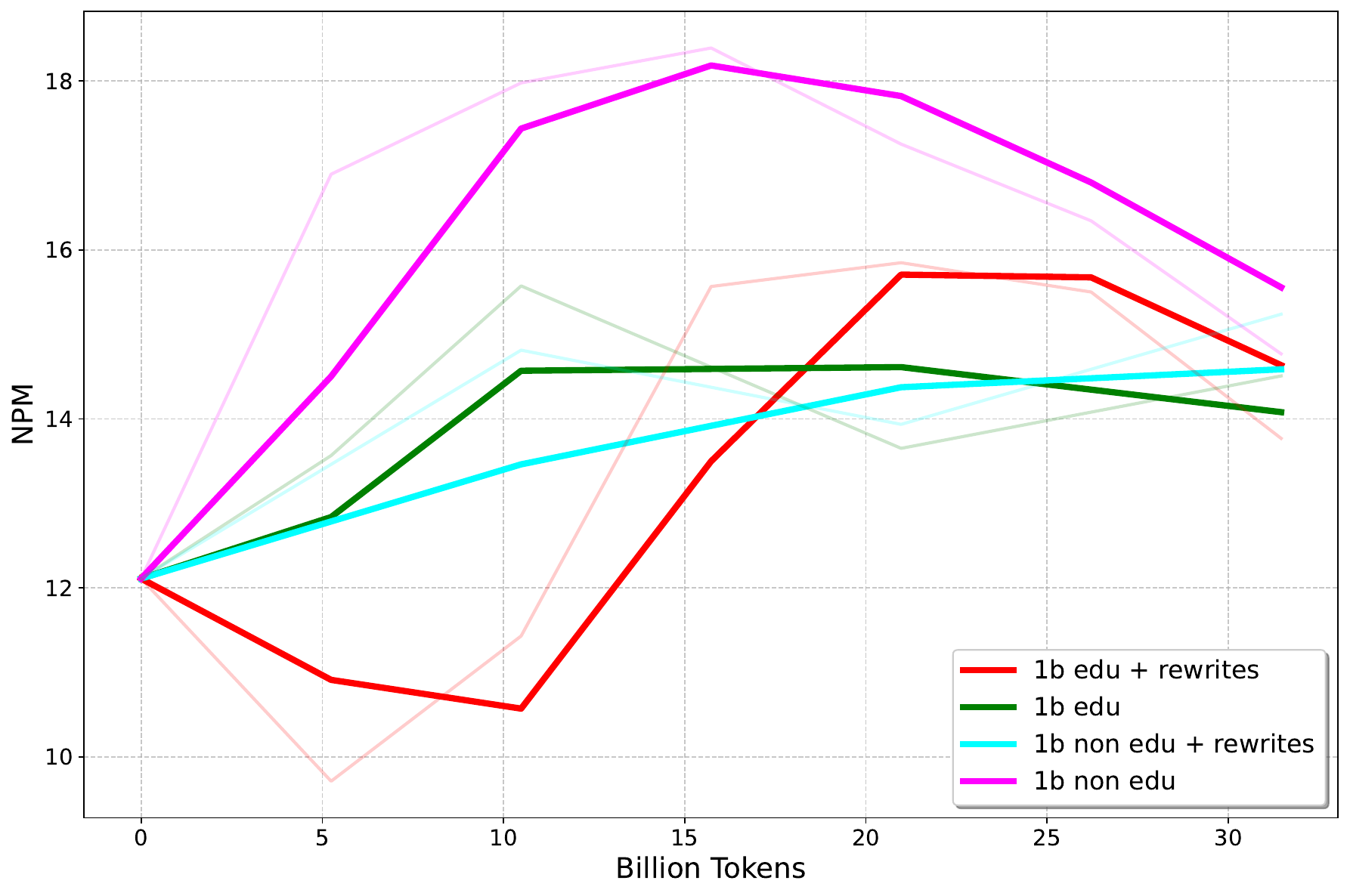}} \hspace*{0.5cm}
\subfloat[Math]{\includegraphics[width=0.31\linewidth]{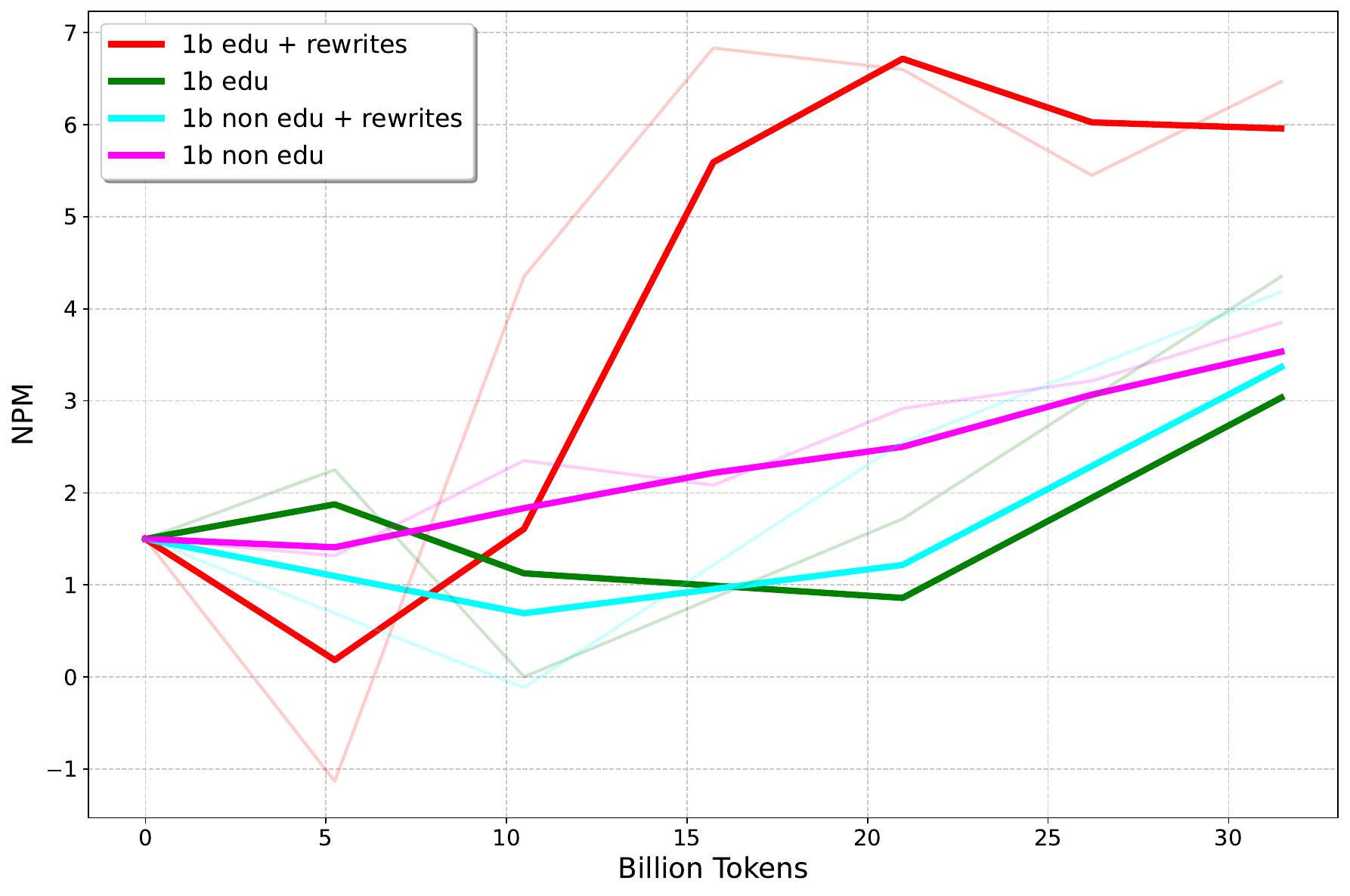}} \\
\subfloat[Social Media]{\includegraphics[width=0.31\linewidth]{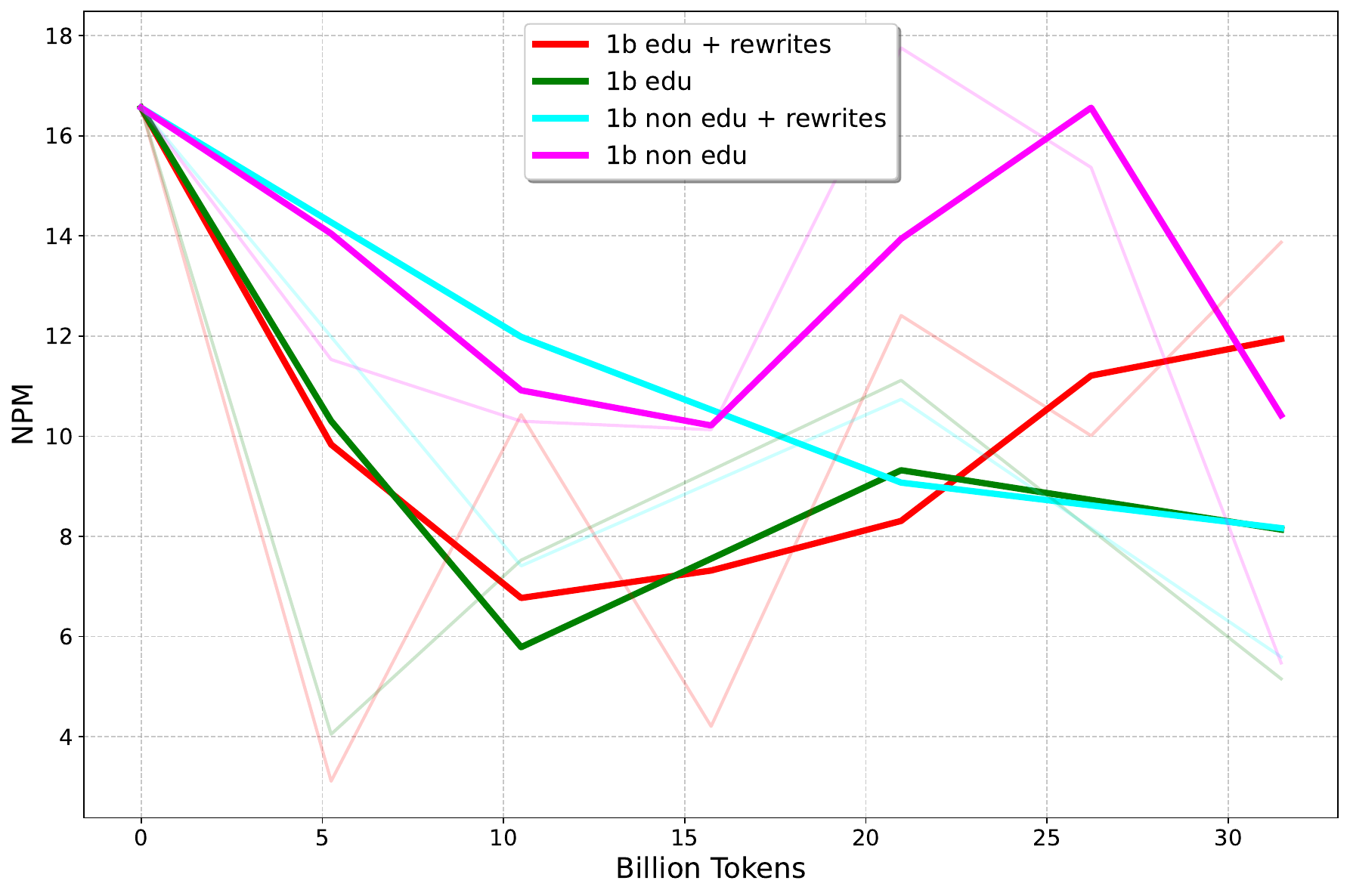}} \hspace*{0.5cm}
\subfloat[General Knowledge]{\includegraphics[width=0.31\linewidth]{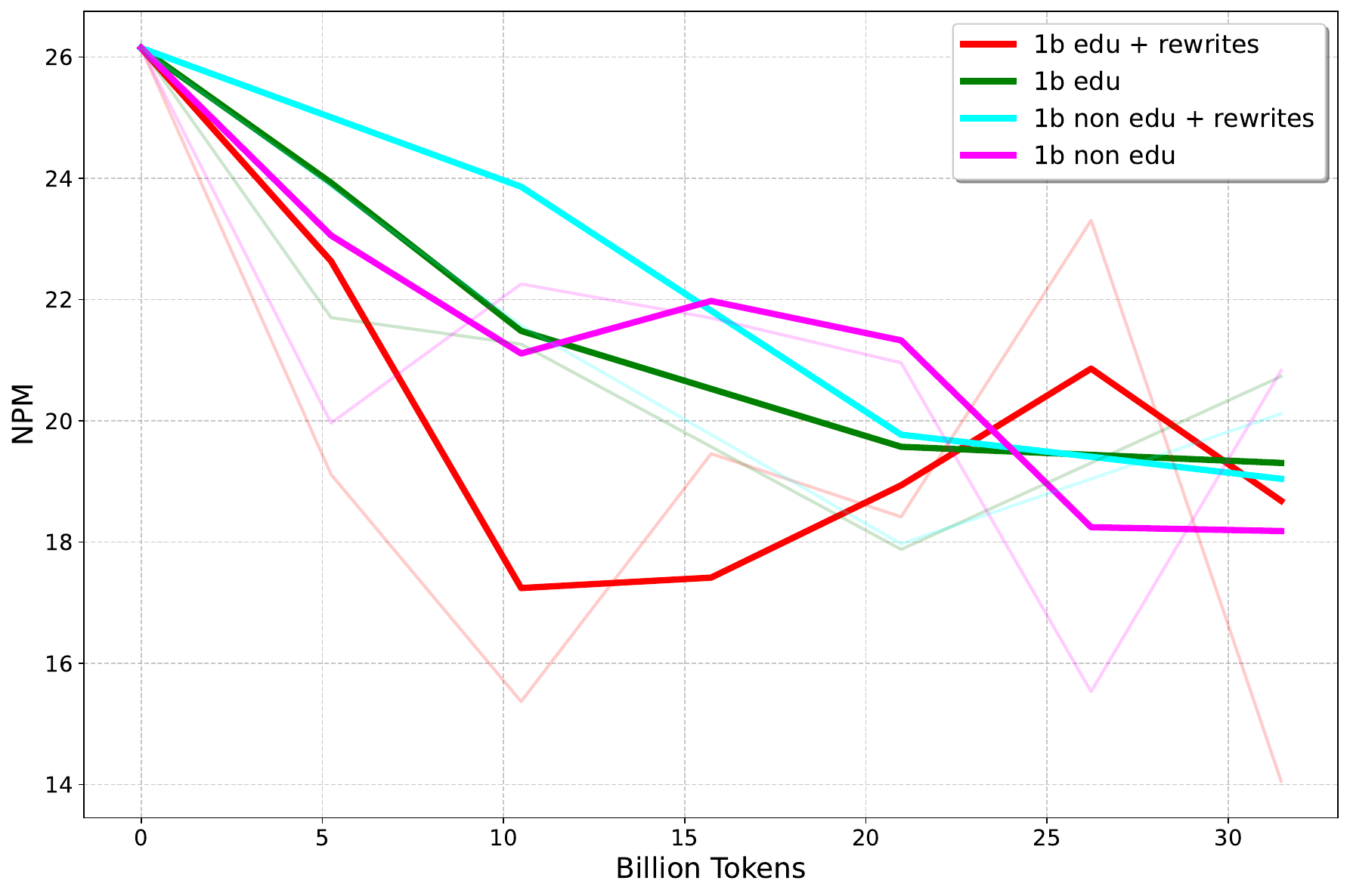}} \hspace*{0.5cm}
\subfloat[Ethics]{\includegraphics[width=0.31\linewidth]{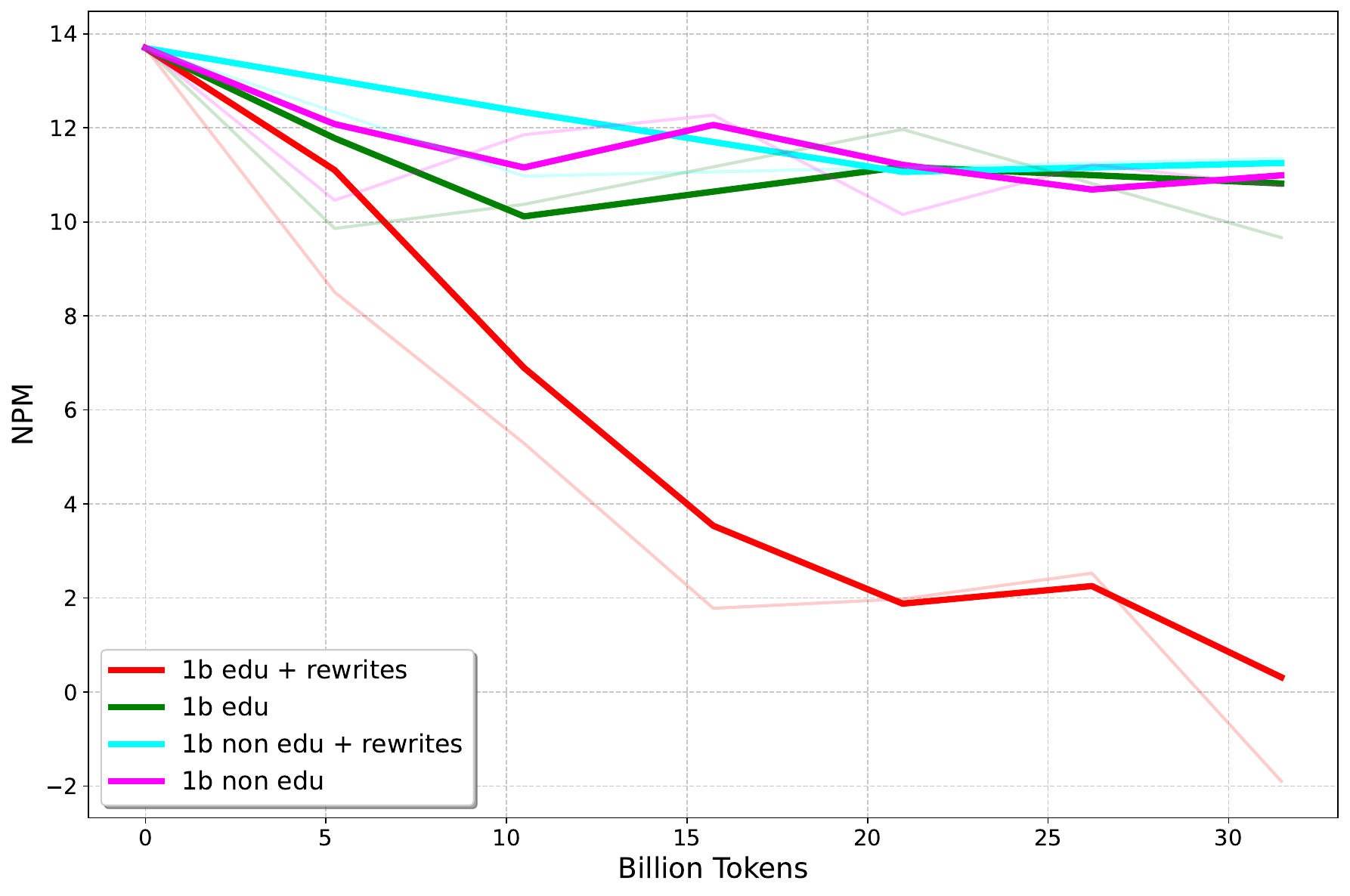}}
\caption{1.1B model: average NPM across task subcategories as a function of training tokens. The nine subcategories shown are those with the largest number of tasks in PoETa~V2.}
\label{fig:1b_subcategories}
\end{figure*}

Figure~\ref{fig:1b_subcategories} shows the subcategory breakdown for the 1.1B model. The patterns are noisier than at the 7B scale, but several trends are visible. The \textit{edu + rewrites} condition tends to lead in extractive QA, sentiment analysis, and classification tasks. However, the \textit{non-edu} condition without rewriting performs well on common sense and general knowledge, consistent with the hypothesis that smaller models benefit from raw data diversity. The overall convergence of conditions in most subcategories suggests that the 1.1B model is less sensitive to data quality and rewriting than the 7B model, further supporting the interpretation that the quality-rewriting interaction is scale-dependent.

\subsection{Summary}

Table~\ref{tab:summary} summarizes the peak NPM achieved by each condition at both model scales. The results confirm that synthetic rewriting and source data quality have a strong positive interaction at the 7B scale, where \textit{edu + rewrites} outperforms all other conditions by a substantial margin. At the 1.1B scale, this interaction is weaker and less consistent, with unmodified low-quality data performing competitively.

\begin{table}[!htb]
\centering
\caption{Peak average NPM on PoETa~V2 for each experimental condition at both model scales. Values in parentheses indicate the number of training tokens (in billions) at which the peak was achieved.}
\label{tab:summary}
\begin{tabular}{lcc}
\hline
\textbf{Condition} & \textbf{7B} & \textbf{1.1B} \\
\hline
Edu + Rewrites     & \textbf{41.0} (30B) & \textbf{15.1} (25B) \\
Edu                & 38.5 (20B) & 13.7 (30B) \\
Non-edu + Rewrites & 35.8 (20B) & 13.8 (30B) \\
Non-edu            & 35.2 (20B) & \textbf{15.1} (20B) \\
\hline
\end{tabular}
\end{table}

\section{Conclusions}

We presented a controlled study of how synthetic document rewriting interacts with source data quality for Portuguese continued pretraining, evaluated across two model scales on 44~tasks from the PoETa~V2 benchmark.

Our main finding is that rewriting acts as a \textbf{quality multiplier}: at the 7B scale, applying style-conditioned rewriting to high-quality data yields a gain of +3.4~NPM over using the same data unmodified, whereas applying the same rewriting procedure to low-quality data provides a marginal improvement of only +0.5~NPM. This asymmetry demonstrates that the benefits of synthetic rewriting are strongly conditioned on the quality of the input data, and that rewriting is most effective when used to amplify already-curated corpora rather than to compensate for poor source quality.

At the 1.1B scale, the quality-rewriting interaction is weaker and less consistent, with unmodified low-quality data performing comparably to rewritten high-quality data. This suggests that the amplification effect is \textbf{scale-dependent}, potentially because smaller models lack sufficient capacity to leverage the structural transformations introduced by rewriting, or because raw data diversity provides a stronger training signal at lower parameter counts.

Our category-level analysis reveals that the quality-rewriting interaction is not uniform across task types. Knowledge-intensive categories such as exams and Brazil-specific tasks show the largest quality effects, while social-media and sentiment tasks are relatively insensitive to both quality and rewriting. Notably, general-knowledge tasks show a case where rewriting can slightly \textit{hurt} performance, suggesting that the rewriting process may occasionally distort factual content.

\textbf{Limitations.} Our results are based on single training runs per configuration, without error bars or statistical significance testing. We use a single rewriting model (Qwen-2.5-7B)---chosen for computational budget reasons---and do not ablate across rewriting model sizes or individual rewriting styles. No tokenizer adaptation was performed, which may limit the efficiency of Portuguese token representation. The rewritten conditions contain more unique tokens than the non-rewritten baselines (approximately 40B vs.\ 10B unique tokens), which partially confounds the comparison; while the asymmetric gain across quality tiers argues against this being the sole explanation, future work should include matched-diversity controls. Additionally, our training mixes rewritten and original data rather than exploring optimal mixing ratios, and the evaluation is restricted to a single benchmark (PoETa~V2).

\textbf{Future work.} Several directions remain open: (i)~ablating individual rewriting styles to identify which transformations contribute most; (ii)~varying the rewriting model size to study whether stronger rewriters produce better training data; (iii)~systematically varying the ratio of rewritten to original data to find optimal mixing strategies; (iv)~adding matched-diversity controls where non-rewritten data is augmented to the same unique token count without rewriting; (v)~extending the study to other low-resource languages to test the generality of the quality-rewriting interaction; and (vi)~performing multiple runs with different seeds to enable proper statistical comparisons.

\bibliography{paper}
\bibliographystyle{apalike}



\end{document}